\documentclass[sigconf]{acmart}
\AtBeginDocument{%
  }

\setcopyright{acmlicensed}

\copyrightyear{2025}
\acmYear{2025}
\setcopyright{acmlicensed}
\acmConference[AIQAM '25]{Proceedings of the 2nd ACM Workshop in AI-powered Question \& Answering Systems}{October 27--28, 2025}{Dublin, Ireland}
\acmBooktitle{Proceedings of the 2nd ACM Workshop in AI-powered Question \& Answering Systems (AIQAM '25), October 27--28, 2025, Dublin, Ireland}
\acmDOI{10.1145/3746274.3760392}
\acmISBN{979-8-4007-2056-7/2025/10}

\usepackage{tcolorbox}

\theoremstyle{definition}
\newtheorem{example}{Example}  

\usepackage{changepage}
\usepackage{algorithm}
\usepackage{algpseudocode}
\usepackage{multirow}
\usepackage{makecell}
\usepackage{booktabs,multirow,array,caption}
\usepackage{longtable}

\usepackage{pifont}  
\usepackage{array} 

\usepackage{dblfloatfix}

\begin{document}

\title{Formal Reasoning for Intelligent QA Systems: A Case Study in the Educational Domain}



\author{Tuan Bui}
\email{tuanbc88@hcmut.edu.vn}
\orcid{0000-0002-8587-182X}
\affiliation{%
  \institution{Ho Chi Minh City University of Technology (HCMUT) - VNU-HCM}
  \city{Ho Chi Minh City}
  \country{VietNam}
}

\author{An Nguyen}
\email{an.nguyen2452023@hcmut.edu.vn}
\orcid{0009-0002-4124-1997}
\affiliation{%
  \institution{Ho Chi Minh City University of Technology (HCMUT) - VNU-HCM}
  \city{Ho Chi Minh City}
  \country{Vietnam}
}

\author{Phat Thai}
\email{phat.thaiquang2004@hcmut.edu.vn}
\orcid{0009-0004-8361-4741}
\affiliation{%
  \institution{Ho Chi Minh City University of Technology (HCMUT) - VNU-HCM}
  \city{Ho Chi Minh City}
  \country{Vietnam}
}

\author{Minh Hua}
\email{minh.hua2006@hcmut.edu.vn}
\orcid{0009-0007-2282-6021}
\affiliation{%
  \institution{Ho Chi Minh City University of Technology (HCMUT) - VNU-HCM}
  \city{Ho Chi Minh City}
  \country{Vietnam}
}

\author{Ngan Pham L.N.}
\email{ngan.pham1409207@hcmut.edu.vn}
\orcid{0009-0007-0963-3495}
\affiliation{%
  \institution{Ho Chi Minh City University of Technology (HCMUT) - VNU-HCM}
  \city{Ho Chi Minh City}
  \country{Vietnam}
}

\author{Ngan Pham T.B.}
\email{ngan.pham11@hcmut.edu.vn}
\orcid{0009-0004-5622-9549}
\affiliation{%
  \institution{Ho Chi Minh City University of Technology (HCMUT) - VNU-HCM}
  \city{Ho Chi Minh City}
  \country{Vietnam}
}

\author{Dung Le}
\email{dung.lengochung@hcmut.edu.vn}
\orcid{0009-0001-9095-0119}
\affiliation{%
  \institution{Ho Chi Minh City University of Technology (HCMUT) - VNU-HCM}
  \city{Ho Chi Minh City}
  \country{Vietnam}
}

\author{Long Nguyen}
\email{long.nguyencse2023@hcmut.edu.vn}
\orcid{0009-0008-7488-4714}
\affiliation{%
  \institution{Ho Chi Minh City University of Technology (HCMUT) - VNU-HCM}
   \city{Ho Chi Minh City}
  \country{VietNam}
}

\author{Thanh-Tung Tran}
\email{tttung@hcmiu.edu.vn}
\orcid{0000-0001-9212-8543}
\affiliation{%
  \institution{International University - VNU-HCM}
  \city{Ho Chi Minh City}
  \country{Vietnam}
}

\author{Thang Bui}
\email{bhthang@hcmut.edu.vn}
\orcid{0000-0001-5021-0549}
\affiliation{%
  \institution{Ho Chi Minh City University of Technology (HCMUT) - VNU-HCM}
  \city{Ho Chi Minh City}
  \country{Vietnam}
}

\author{Tho Quan}
\email{qttho@hcmut.edu.vn}
\orcid{0000-0003-0467-6254}
\affiliation{%
  \institution{Ho Chi Minh City University of Technology (HCMUT) - VNU-HCM}
  \city{Ho Chi Minh City}
  \country{Vietnam}
}

\renewcommand{\shortauthors}{Tuan Bui et al.}

\begin{abstract}

Reasoning is essential for closed-domain QA systems in which procedural correctness and policy compliance are critical. While large language models (LLMs) have shown strong performance on many reasoning tasks, recent work reveals that their reasoning traces are often unfaithful—serving more as plausible justifications than as causally grounded derivations. Efforts to combine LLMs with symbolic engines (e.g., Prover9, Z3) have improved reliability but remain limited to static forms of logic, struggling with dynamic, state-based reasoning such as multi-step progressions and conditional transitions.

In this paper, we propose \textbf{MCFR} (Model Checking for Formal Reasoning), a neuro-symbolic framework that integrates LLMs with model checking to support property verification. MCFR translates natural language into formal specifications and verifies them over transition models. To support evaluation, we introduce \textbf{EduMC-QA}, a benchmark dataset grounded in real academic procedures. Our results show that MCFR improves reasoning faithfulness and interpretability, offering a viable path toward verifiable QA in high-stakes closed-domain applications. In addition to evaluating MCFR, we compare its performance with state-of-the-art LLMs such as ChatGPT, DeepSeek, and Claude to contextualize its effectiveness.

\end{abstract}




\received{11 July 2025}

\maketitle

\section{Introduction}

Question Answering (QA) has emerged as a central task in natural language understanding (NLU) due to its wide applicability in domains such as education, healthcare, and human-computer interaction. A key aspect of QA lies in its inherent requirement for reasoning—drawing logical conclusions from given premises or background knowledge. While traditional information retrieval-based QA systems focus primarily on pattern matching or keyword alignment, modern intelligent QA systems aim to reason about input questions using deeper semantic understanding. This work is presented as part of the AIQAM'25 workshop on AI-powered Question Answering Systems for Multimedia, which highlights emerging approaches in intelligent QA across domains and modalities \cite{aiqam25}.

Recent advances in large language models (LLMs) have significantly enhanced QA performance, particularly in tasks involving mathematical or symbolic reasoning. Techniques like Chain-of-Thought (CoT) prompting \cite{kojima2022large} and Multi-Agent Reasoning \cite{miaoselfcheck} simulate human-style step-by-step thinking and have achieved impressive results. However, growing evidence indicates that LLMs often fail at \textit{faithful reasoning}—i.e., their generated reasoning traces do not causally influence the final answer \cite{turpin2023language, maoneuro}. These models tend to produce post hoc rationalizations rather than genuinely grounded inferences, exposing a gap between surface-level coherence and internal logical validity. To close this gap, neuro-symbolic (NeSy) systems offer a promising hybrid approach by combining the representational power of LLMs with the rigor of symbolic solvers. In this paradigm, LLMs translate NL inputs into formal representations (e.g., first-order logic), which are then processed by engines like Z3 or Prover9 \cite{pan2023logiclm}. 

\begin{tcolorbox}[colback=gray!5, colframe=black!40, boxrule=0.5pt, arc=2mm, left=1mm, right=1mm, top=1mm, bottom=1mm]
\begin{example}\label{ex:student-lifecycle}
\textbf{Questions on Student Lifecycle.} \\
\textbf{Context.} The academic progression of a student typically involves multiple stages, such as enrollment, advancement through academic years (Year 1–4), participation in courses and social activities, taking temporary leave, and ultimately reaching either graduation or program withdrawal. Certain actions, such as course registration, are constrained by institutional rules (e.g., prerequisite requirements), while others are subject to timing and eligibility conditions.

\textbf{Questions.} The following questions illustrate four core verification properties that are critical in ensuring the correctness of such educational processes: 
\begin{adjustwidth}{0.5em}{0pt}
$\diamond$ \textbf{Q1 (Liveness).} Can a student graduate within eight semesters if all requirements are satisfied on time? \\
$\diamond$ \textbf{Q2 (Safety).} Is it possible to register for CourseX without passing its prerequisite CourseY? \\
$\diamond$ \textbf{Q3 (Reachability).} Can a student advance to Year 3 within three semesters?\\
$\diamond$ \textbf{Q4 (Fairness).} Could a student accumulate 100 credits yet fail to advance to Year 3 owing to missing social activity points?
\end{adjustwidth}
\end{example}
\end{tcolorbox}

However, symbolic solvers are typically limited to static logic tasks and struggle with dynamic, state-dependent reasoning---\linebreak common in domains like education, where actions such as course registration or academic progression depend on temporal constraints and evolving eligibility. As illustrated in Example~\ref{ex:student-lifecycle}, questions involving safety, liveness, reachability, and fairness require reasoning over sequential transitions and conditional policies—posing challenges for symbolic solvers due to limited procedural expressiveness, and for LLMs due to a lack of verifiable grounding. 

For instance, Q1 (Liveness) asks whether a student can graduate within eight semesters if all requirements are met on time, requiring that a ``good'' state is eventually reachable. In contrast, Q2 (Safety) checks whether the system prevents invalid paths, such as registering for a course without completing its prerequisite. Q3 (Reachability) concerns whether an accelerated academic path is viable—for example, if a student can reach Year 3 in just three semesters by fulfilling requirements early. Q4 (Fairness) explores whether non-academic factors, like missing social activity points, could unfairly block academic advancement despite strong academic performance. Both cases underscore the need for nuanced temporal and policy-aware reasoning, which remains a challenge for current methods.

To address these limitations, we introduce \textbf{MCFR} (Model Checking for Formal Reasoning), a hybrid framework that augments QA systems with verifiable inference capabilities. MCFR employs LLMs to generate formal representations from NL input, which are then verified using model checking (MC) over full state-transition systems. This allows the system to capture procedural behaviors and validate complex properties such as reachability or fairness in academic workflows. To facilitate systematic evaluation, we also construct a new dataset grounded in real-world educational processes that require both symbolic abstraction and temporal reasoning.

\noindent Our contributions are as follows.
\begin{itemize}
\item We propose \textbf{MCFR}, a formal reasoning framework for QA that integrates LLM-based semantic translation with model checking for verifying temporal and procedural properties.
\item We implement a QA prototype grounded in real academic scenarios at Ho Chi Minh City University of Technology (HCMUT), demonstrating the feasibility of state-based formal verification in educational domains.
\item We construct a novel QA dataset, \textbf{EduMC-QA}, based on real institutional processes at HCMUT, and use it to perform comprehensive evaluations assessing the reasoning capability, coverage, and faithfulness of MCFR. We also compare MCFR’s performance against state-of-the-art LLMs such as ChatGPT\footnote{\url{https://chatgpt.com}}, Claude\footnote{\url{https://claude.ai}}, and DeepSeek\footnote{\url{https://chat.deepseek.com}} to contextualize its effectiveness.

\end{itemize}


\section{Related Works}
\subsection{Reasoning Patterns of LLMs}
LLMs have demonstrated strong performance across various reasoning tasks, particularly through prompting-based techniques such as CoT prompting~\cite{kojima2022large}, along with various enhancements that elicit intermediate reasoning steps and improve answer accuracy~\cite{liu-etal-2025-logic}. Recent advances also explore multi-agent reasoning setups, in which one LLM generates the reasoning steps while another acts as a verifier, identifying errors and prompting the first model to revise its reasoning accordingly~\cite{miaoselfcheck}.

Despite these improvements, concerns remain about the faithfulness of LLM-generated reasoning. Studies have shown that models often produce coherent reasoning traces that are not causally tied to their final answers. For example, \cite{turpin2023language} inserted misleading or biased cues into reasoning chains and found that LLMs often relied on these biases rather than engaging in true inference. Similarly, \cite{matton2024walk} manipulated reasoning steps—by removing key inferences or injecting noise—and observed that LLM predictions remained mostly unchanged. These findings suggest that current LLMs frequently treat reasoning traces as post hoc justifications rather than integral components of decision-making, raising critical questions about their reliability in high-stakes applications.

\subsection{Formal Reasoning}  

Recent work has explored integrating LLMs with symbolic solvers to achieve more faithful and verifiable reasoning. Logic-LM \cite{pan2023logiclm} decomposes the reasoning process into three phases: problem formulation, symbolic inference, and interpretation—showing improvements over traditional CoT-style prompting. Other approaches such as SpecGen \cite{ma2024specgen} use mutation-based corrections and verifier feedback to iteratively refine LLM-generated formal specifications. Logic-guided theorem proving \cite{pan2023logiclm} further scaffolds LLM outputs with formal logic to reduce semantic errors. In practical applications, LLMs have also been combined with SMT solvers for task planning \cite{hao2025llm}, where the solver verifies constraints and offers corrective feedback during planning failures.

Despite these advances, symbolic solvers like Z3 or Prover9 remain inherently static: they require precise and well-formed logical inputs and struggle with ambiguity, incomplete specifications, or procedural reasoning involving state changes over time. This rigidity makes them fragile in QA pipelines where LLM outputs may be noisy or underspecified. Furthermore, such solvers lack mechanisms to reason over dynamic transitions or evolving states—common in domains like education—limiting their standalone applicability without iterative repair or hybrid integration.

These limitations motivate the development of hybrid frameworks like MCFR, which combine LLM-based translation with state-based formal verification techniques such as model checking.

\section{Background}

\subsection{Neural Symbolic AI}

NeSy systems aim to combine the strengths of neural networks—such as perception and natural language understanding—with the formal rigor and interpretability of symbolic reasoning. A common architecture involves using neural models (e.g., LLMs) to translate natural language inputs into formal representations like first-order logic, which are then passed to symbolic engines (e.g., theorem provers or model checkers) for inference. This modular design allows LLMs to handle linguistic ambiguity while relying on symbolic solvers for precise, verifiable reasoning \cite{maoneuro}.

Recent surveys, such as \cite{10721277}, have categorized NeSy systems into various types based on the level of integration between neural and symbolic components. In many practical QA applications, loosely coupled hybrids—where LLMs act as semantic translators and symbolic modules verify logical constraints—have shown promising results.

\subsection{Model Checking}

Model checking \cite{10.1007/BFb0058022} is an automated formal verification technique used to determine whether a finite-state system model satisfies a given logical specification. The system is typically modeled as a transition system $M$, and the specification $\varphi$ is expressed in Computation Tree Logic (CTL). CTL enables the formal specification of critical system properties, including safety (the avoidance of undesirable states), liveness (the guarantee of system progress), reachability (the ability to attain desired goal states), and fairness (the assurance that no enabled action is indefinitely postponed).

Given a transition system and a CTL formula, model checking tools apply algorithms to exhaustively explore the system's state space, thereby enabling the formal verification of the specified properties. This makes model checking a robust and effective method for reasoning about procedural behaviors across various domains, including educational systems.

\section{MCFR Framework}

\subsection{Overall Architecture}

\begin{figure}[!h]
    \centering
    \includegraphics[width=0.99\linewidth]{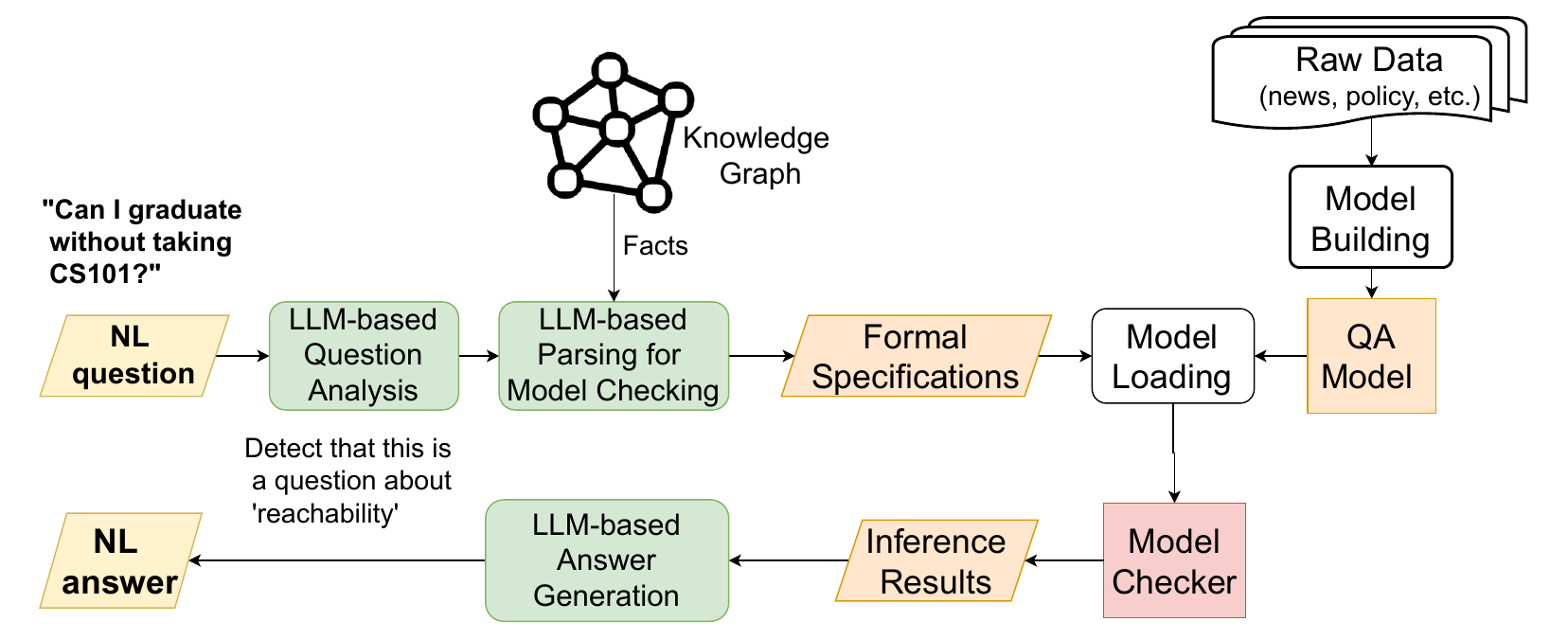}
    \caption{Overview of the MCFR framework integrating LLMs with formal verification via model checking.}
    \label{fig:overral_arch_qa}
\end{figure}

Our proposed framework comprises four main components that collaboratively enable question answering grounded in formal verification. The architecture is illustrated in Figure~\ref{fig:overral_arch_qa}.

\paragraph{(1) Model Building Phase} 
In the first stage, domain experts manually analyze raw data sources such as policy documents, institutional guidelines, and official announcements. These unstructured texts are used to construct a domain-specific QA model that encodes the system behavior, logical constraints, and domain rules required for formal reasoning. This process is performed offline and serves as the static knowledge base for the downstream inference pipeline.  
The detailed implementation of this QA model using a concrete model checking tool is presented in Section~\ref{sec:formal-model-impl}.

\paragraph{(2) Question Processing and Specification Generation}
Given a user-issued NL question, the system first performs LLM-based Question Analysis to determine the intent of the query (e.g., identifying whether the question pertains to ``reachability''). The system then queries the Knowledge Graph (Appendix \ref{app:edu_kg}) to retrieve relevant facts associated with the question context. These inputs are processed by a LLM-based Parsing component that translates the question and facts into formal specifications compatible with model checking formalisms.

\begin{figure*}[t]
    \centering
    \includegraphics[width=0.95\linewidth]{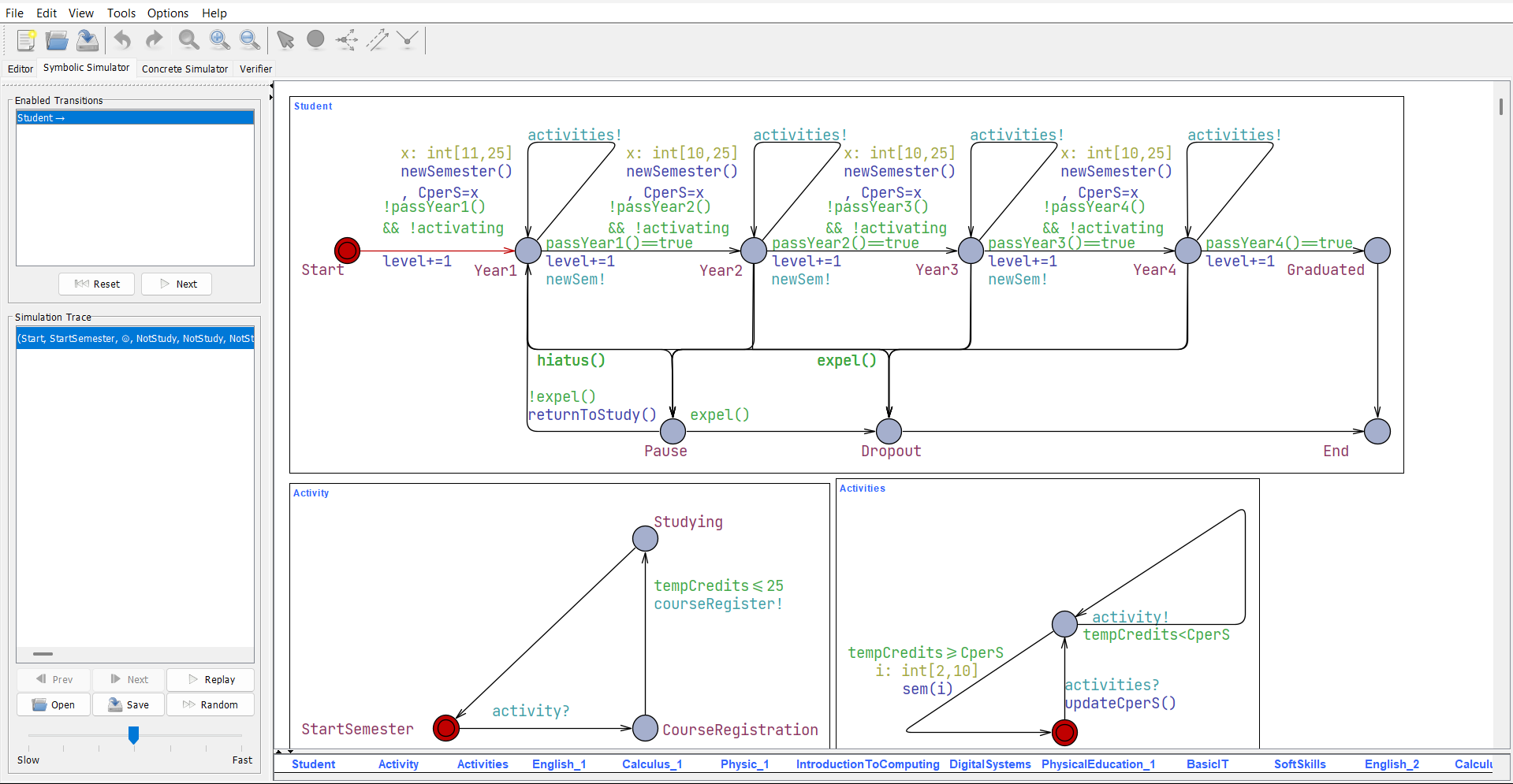}
    \caption{UPPAAL implementation of the student lifecycle model.}
    \label{fig:uppaal-student-model}
\end{figure*}

\begin{table*}[!t]
\small
\centering
\renewcommand{\arraystretch}{1.2}
\begin{tabular}{|p{4.5cm}|p{3.5cm}|p{1.0cm}|p{6.5cm}|}
\hline
\textbf{Natural–language Questions} & \textbf{UPPAAL Query} & \textbf{Answer} & \textbf{Representative Trace} \\ \hline
\textit{Q1.} Can a student graduate within eight semesters if all requirements are satisfied on time? 
& \texttt{E<> (Student.Graduated \&\& time $\le$ 8)} 
& \textbf{Yes} 
& \texttt{Start} $\rightarrow$ \texttt{Year1} $\rightarrow$ \texttt{Year2} $\rightarrow$ \texttt{Year3} $\rightarrow$ \texttt{Year4} $\rightarrow$ \texttt{Graduated}; \texttt{time}=8, \texttt{totalCredits}=128. \\ \hline

\textit{Q2.} Is it possible to register for \textsf{CourseX} without passing its prerequisite \textsf{CourseY}?
& \texttt{E<> (CourseX\_Reg \&\& !CourseY\_Passed)} 
& \textbf{No} 
& --- (no path exists; the guard \texttt{CourseY\_Passed} on the \textsf{CourseX} transition blocks all attempts). \\ \hline

\textit{Q3.} Can a student advance to Year 3 within three semesters?
& \texttt{E<> (Student.Graduated \&\& time $\le$ 3)}
& \textbf{No} 
&
--- (no path exist; completing \texttt{English\_4} takes at least four semesters)
\\ \hline

\textit{Q4.} Could a student accumulate 100 credits yet fail to advance to Year~3 owing to missing social activity points?
& \texttt{E<> (totalCredits $\ge$ 100 \&\& Student.Year4} 
& \textbf{Yes} 
& \texttt{Start} $\rightarrow$ \texttt{Year1} $\rightarrow$ \texttt{Year2}; take electives until \texttt{totalCredits}>=100, but keep \texttt{passEnglish\_4}$==false$; no transition to Year~3 enabled. \\ \hline
\end{tabular}
\caption{Mapping from NL questions to UPPAAL queries, with model‐checking outcomes and illustrative execution paths.}
\label{tab:nl-to-uppaal}
\end{table*}

\paragraph{(3) Model Checking Inference} 
The generated formal specifications and the pre-built QA model are passed to the Model Loading component, which prepares them for verification. These artifacts are then submitted to a Model Checker, which performs symbolic reasoning and evaluates whether the specifications hold over the model. The outcome is returned as structured Inference Results, containing formally validated information relevant to the user query.

\paragraph{(4) Natural Language Answer Generation}
Finally, the LLM-based Answer Generation module converts the inference results into a coherent natural language answer. This ensures that users receive not only logically sound but also linguistically accessible responses.

\subsection{MCFR Implementation}
\label{sec:formal-model-impl}

Our MCFR framework follows the \textit{NeSy Type~2} paradigm, integrating neural and symbolic components in a sequential pipeline. Specifically, Components~2 and~4 leverage the Google Gemini model\footnote{\url{https://deepmind.google/technologies/gemini/}} for natural language processing and high-level conversion tasks. In contrast, Components~3 and~4 focus on formal verification, where the domain model is constructed and executed using the UPPAAL\footnote{\url{https://uppaal.org}} tool. This combination allows MCFR to bridge free-form natural language understanding with precise, state-based reasoning, ensuring both expressiveness and verifiability.

Building on the student lifecycle scenario in Example~\ref{ex:student-lifecycle}, we construct a formal model using UPPAAL, where the academic progression of a student is encoded as a network of interacting timed automata. Each component captures distinct behaviors such as course registration, semester-level coordination, and academic status transitions, all governed by temporal constraints and institutional rules (e.g., prerequisites, credit thresholds). This executable model enables simulation and formal verification of key properties such as \textit{safety} (e.g., preventing invalid registration), \textit{liveness} (e.g., ensuring eventual graduation), and \textit{reachability} (e.g., returning from academic leave). These capabilities form the foundation for answering structured queries in our dataset.

The model uses global state variables like \texttt{totalCredits}, \texttt{time}, and Boolean flags (e.g., \texttt{passEnglish\_2}, \texttt{passBasicIT}) to represent student status. Year-level progression is constrained by guard functions (e.g., \texttt{passYear1()}, \texttt{passYear2()}) that check eligibility conditions. These guards are attached to transitions to ensure policy compliance. Figure~\ref{fig:uppaal-student-model} illustrates the full model structure, while Table~\ref{tab:nl-to-uppaal} presents sample verification questions along with their corresponding formal properties and execution traces.

\section{EduMC-QA Dataset}

To evaluate the capabilities of QA systems in performing verifiable reasoning, we construct a novel dataset, \textbf{EduMC-QA}, centered on procedural and temporal logic within the educational domain. Rather than focusing on factual recall, the questions in EduMC-QA are designed to assess whether certain properties hold over dynamic systems, such as academic workflows. Each question is annotated with one of four key verification types inspired by formal methods, which reflect core reasoning challenges commonly addressed in model checking.

\vspace{0.5em}
\noindent\textbf{Annotation Guidelines.}
Each question in EduMC-QA is manually assigned to one of the four verification categories based on its semantic intent and the property it seeks to verify within the academic system. The classification is grounded in definitions widely used in formal verification, with specific identification cues as shown in Table~\ref{tab:classification-guidelines}.

\begin{table}[!h]
\centering
\small
\caption{Verification types with classification criteria}
\label{tab:classification-guidelines}
\begin{tabular}{p{1.2cm}p{6.5cm}}
\toprule
\textbf{Type} & \textbf{Classification Criteria} \\
\midrule
Safety & $\diamond$ Ensures that nothing bad ever happens. \newline $\diamond$ Checks whether the system prevents invalid or undesirable actions (e.g., bypassing prerequisites). \\
\addlinespace
Liveness & $\diamond$ Verifies that something good will eventually happen. \newline $\diamond$ Asks whether a desirable outcome can be achieved assuming all conditions are met (e.g., graduation). \\
\addlinespace
Reachability & $\diamond$ Asks whether a specific state can be reached from the current state. \newline $\diamond$ Considers whether a goal can be achieved under certain conditions (e.g., fast-tracking to Year 3). \\
\addlinespace
Fairness & $\diamond$ Evaluates whether the system behaves in a just or non-discriminatory way. \newline $\diamond$ Checks for unreasonable blocking despite strong compliance (e.g., enough credits but still blocked). \\
\bottomrule
\end{tabular}
\end{table}

To ensure annotation quality and conceptual alignment, each question was independently categorized by three domain experts with backgrounds in formal methods and logical reasoning. The final label was determined by majority vote, and in the rare case of full disagreement, resolution was achieved through structured discussion and consensus.

\vspace{0.5em}
\noindent\textbf{Illustrative Classification Examples.}
For instance, the question Q1-``Can a student graduate within eight semesters if all requirements are satisfied on time?'' is labeled as \emph{Liveness} because it asks whether a desirable end state (graduation) will eventually occur given favorable conditions, emphasizing guaranteed progress rather than rule enforcement. Conversely, Q2-``Is it possible to register for CourseX without passing its prerequisite CourseY?'' is labeled as \emph{Safety} since it seeks to prevent an invalid action (bypassing a prerequisite), thus ensuring that no undesirable state can occur in the system.

\vspace{0.5em}
\noindent\textbf{Example Item.}  
Each item in the dataset consists of a natural language question, a formal model, a specification, and a verification result. Additionally, two auxiliary fields are included: \texttt{context}, which contains relevant logical facts extracted from the knowledge base, and \texttt{formal\_spec}, which represents the parsed formal specification and contextual information expressed in UPPAAL syntax. A simplified example is shown below:

\begin{quote}
\ttfamily
\small
\{\\
\ \ "q": "Can a student enroll in a course without completing its prerequisite?",\\
\ \ "cat": "Safety",\\
\ \ "context": [ "prerequisite(CourseX, CourseY)",\\
\ \ \ \ "student\_has\_not\_completed(CourseY)" ],\\
\ \ "spec": "AG (enroll\_request -> completed(prereq))",\\
\ \ "formal\_spec": "\\texttt{A[] (enroll\_request imply completed[prereq])}",\\
\ \ "groundtruth": "No"\\
\}
\end{quote}

\begin{table}[h]
\centering
\small
\caption{Question distribution by verification category in EduMC-QA}
\label{tab:dataset-stats}
\begin{tabular}{p{1.5cm}p{0.7cm}p{5.0cm}}
\toprule
\textbf{Verification Property} & \textbf{\#Q} & \textbf{Example Type} \\
\midrule
Safety        & 34 & Invalid course registration \\
Liveness      & 34 & Progress toward internship / graduation \\
Reachability  & 38 & Path to graduation from Year 1 \\
Fairness      & 23 & LOA approval and return tracking \\
\midrule
\textbf{Total} & 129 & --- \\
\bottomrule
\end{tabular}
\end{table}

\noindent\textbf{Dataset Statistics.}  
EduMC-QA comprises 129 natural language questions, each paired with a formal verification instance. The distribution across the four verification categories (Table~\ref{tab:dataset-stats}) ranges from 23 to 38 items per category. The questions span diverse academic scenarios—such as enrollment, graduation, leave tracking, and internships—and vary in both linguistic and logical complexity. All items are manually annotated and validated for semantic correctness.


\section{Experimentation}


\begin{table*}[!t]
\centering
\small
\caption{Representative verification questions and comparison of reasoning accuracy across MCFR and LLM baselines}
\label{tab:llm-comparison-showcase}
\renewcommand{\arraystretch}{1.25}
\begin{tabular}{|p{5.0cm}|p{1.1cm}|p{1.0cm}|p{0.8cm}|p{0.9cm}|p{0.8cm}|p{5.0cm}|}
\hline
\textbf{Natural-language Question} & \textbf{Ground Truth} & \textbf{MCFR (Ours)} & \textbf{Chat GPT} & \textbf{Claude} & \textbf{Deep Seek} & \textbf{Observation / Reasoning Error} \\
\hline

\textbf{Q1 (Liveness).} Can a student graduate within eight semesters if all requirements are satisfied on time? & 
\textbf{Yes} & \ding{51} & \ding{51} & \ding{55} & \ding{51} & 
Claude need further data to conclude \\

\hline

\textbf{Q2 (Safety).} Is it possible to register for CourseX without passing its prerequisite CourseY? & 
\textbf{No} &  \ding{51} & 
\ding{55} & 
\ding{55} & 
\ding{55} & 
All models fail to enforce prerequisite constraints; misinterpret the question as policy-based. \\

\hline

\textbf{Q3 (Reachability).} Can a student advance to Year 3 within three semesters? & 
\textbf{No} &  \ding{55} & 
\ding{51} & 
\ding{51} & 
\ding{51} & 
All models ignore sequential prerequisites, while completing \texttt{English\_4} requires at least four semesters.
 \\

\hline

\textbf{Q4 (Fairness).} Could a student accumulate 100 credits yet fail to advance to Year 3 owing to missing social activity points? & 
\textbf{Yes} &  \ding{51} & 
\ding{55} & 
\ding{55} & 
\ding{55} & 
The models do not correctly infer the academic rule that promotion to the third year requires the completion of \texttt{English\_4}. \\

\hline
\end{tabular}
\end{table*}

Building on the implementation introduced in earlier sections, we now evaluate the effectiveness of our MCFR framework by comparing its verification outcomes against responses from state-of-the-art LLMs. Using a representative academic scenario modeled in UPPAAL, we assess how well LLMs handle formal reasoning tasks and analyze their strengths and limitations in contrast to our model checking approach.

\subsection{Case Study: Reasoning Comparison}

To better illustrate the limitations of current LLMs in handling formal verification tasks, we present a qualitative case study comparing our neuro-symbolic system (\textbf{MCFR}) with state-of-the-art LLMs: ChatGPT, Claude, and DeepSeek. We select one representative question from each of the four verification categories in EduMC-QA—Safety, Liveness, Reachability, and Fairness—and examine whether these models can correctly answer questions that require logical inference over structured academic rules.

Each case is based on a prompt grounded in formal academic regulations, as illustrated in Example~\ref{ex:promt-input-llms} (a summarized version of the full academic policy; the complete context is available at~\footnote{\url{https://github.com/tuanbc88-hcmut/mcfr_modelchecking_qa.git}}). For instance, a liveness question asks whether a student can graduate within eight semesters if all requirements are satisfied on time. A safety question examines the possibility of course registration without completing prerequisites. These tasks demand reasoning over sequential transitions and conditional rules—capabilities that remain challenging for both LLMs (due to lack of verifiable grounding) and symbolic solvers (due to limited procedural expressiveness).


\begin{tcolorbox}[colback=gray!5, colframe=black!40, boxrule=0.5pt, arc=2mm, left=1mm, right=1mm, top=1mm, bottom=1mm]
\small
\begin{example}\label{ex:promt-input-llms}

\textbf{Prompt Input for Evaluating LLM Reasoning}

\textbf{Context.} Students may retake failed courses (no credit earned until passed) or improve grades by retaking passed courses with the same code (highest score used for GPA; credits unchanged). To qualify as a 2nd- or 3rd-year student, one must meet credit, language, IT, and other institutional requirements (e.g., defense education). Additionally, off-campus internships require at least 65\% of total credits and must precede the graduation thesis, which also requires a completed specialized project.
(See full example at:~\footnote{\url{https://github.com/tuanbc88-hcmut/mcfr_modelchecking_qa.git}})

\textbf{Task.} Based on the context above, answer the four verification questions presented in Example~\ref{ex:student-lifecycle}. Each answer should be \textit{Yes}, \textit{No}, or \textit{Uncertain}, and must be accompanied by a justification.
\end{example}
\end{tcolorbox}

\begin{table}[t]
\centering
\small
\caption{Reasoning accuracy (\%) by question type and models}
\label{tab:reasoning-performance}
\begin{tabular}{
p{2.0cm}
>{\raggedleft\arraybackslash}p{0.6cm}
>{\raggedleft\arraybackslash}p{0.8cm}
>{\raggedleft\arraybackslash}p{1.3cm}
>{\raggedleft\arraybackslash}p{0.8cm}
>{\raggedleft\arraybackslash}p{0.7cm}}
\toprule
\textbf{Model} & \textbf{Safety (34)} & \textbf{Liveness (34)} & \textbf{Reachability (38)} & \textbf{Fairness (24)} & \textbf{Total (130)} \\
\midrule
ChatGPT-casual & 79.41 & 82.35 & 63.16 & 66.67 & 73.08 \\
ChatGPT-reason & 73.53 & 88.24 & 84.21 & 83.33 & 82.31 \\
\addlinespace[0.35em]
Claude-Sonnet & 44.12 & 50.00 & 44.74 & 12.50 & 40.00 \\
\addlinespace[0.35em]
DeepSeek-casual & 76.47 & 79.41 & 76.32 & 75.00 & 76.92 \\
DeepSeek-R1 & 52.94 & 55.88 & 50.00 & 37.50 & 50.00 \\
\addlinespace[0.35em]
\textbf{MCFR(Ours)} & 94.12 & 97.06 & 94.74 & 87.50 & 93.85 \\
\bottomrule
\end{tabular}
\end{table}

Table~\ref{tab:llm-comparison-showcase} summarizes the correctness of model responses, using \ding{51} for correct and \ding{55} for incorrect answers. Table~\ref{tab:reasoning-performance} further compares the overall reasoning accuracy across models on the EduMC-QA dataset. Detailed model responses are provided in Appendix~\ref{app:llms_response}.

\subsection{Discussion}
Among LLMs, ChatGPT-reason performs best overall, particularly in liveness and reachability questions, suggesting a certain ability to reason over sequential constraints. DeepSeek-casual also delivers consistent results, while Claude-Sonnet underperforms significantly---especially on fairness---highlighting its difficulty with multi-step, policy-based reasoning. In contrast, MCFR achieves reliable reasoning via formal verification. However, it relies on a manually constructed domain model, an expert-driven and time-consuming process, which limits scalability. This points to future potential in semi-automated model construction, where LLMs assist in abstraction and experts validate critical elements. Another key challenge lies in translating NL into formal specifications; errors in this step can lead to incorrect verification outcomes even when the model is sound. Robust validation mechanisms are therefore essential to ensure end-to-end QA integrity.

\noindent\textbf{NL-to-CTL Conversion Errors.} A major source of error arises in converting natural language questions into formal UPPAAL queries. Current LLMs lack knowledge of the UPPAAL syntax, making it difficult to map NL descriptions into correct CTL formulas without explicit information about the target model. For accurate NL$\rightarrow$CTL translation, the LLM must be aware of both the UPPAAL model in use and its precise syntactic rules. Moreover, there are currently no established metrics for systematically measuring the quality of NL-to-CTL conversion, which makes it harder to benchmark and improve this step. We argue that improving the prompt---rather than extending the underlying logic model---is a more effective and immediate solution. This is because UPPAAL requires absolute syntactic correctness for verification; while logic model enhancements may improve runtime efficiency, they do not inherently boost accuracy. In contrast, prompt refinement can guide the LLM toward producing syntactically valid and semantically aligned queries. 

For example, given the NL question: \textit{``Can a student graduate within eight semesters if all requirements are satisfied on time?''}, the correct query is \texttt{E<> (Student.Graduated \&\& time $\leq$ 8)}. Without prompt tuning, LLM might output \texttt{E<> (semester <= 8 \&\& AllRequirementsMet \&\& Student.Graduated)}, which is incorrect for two reasons: (1) the model uses the variable \texttt{time} rather than \texttt{semester}, and (2) \texttt{AllRequirementsMet} does not exist in the UPPAAL model.

\noindent\textbf{QA Model Scalability and Domain Adaptation.} Constructing and validating the current QA model required approximately 56 person-hours. Reproducing the model for other universities or organizations would take about 2--8 hours, depending on the extent of policy differences. The process could follow three main steps:  
(1) Modify institutional rules affecting student progression and graduation (up to 8 hours depending on complexity),  
(2) Adjust parameter values such as minimum passing grades, and  
(3) Conduct thorough verification testing.  
Beyond the academic domain, the approach can be adapted to other fields such as law or healthcare. In such cases, the time and complexity of adaptation depend heavily on the number of states, transitions, and governing rules that must be identified and modeled within each specific domain.

\section{Conclusion}
We introduced MCFR, a neuro-symbolic QA framework that integrates LLM-based parsing with model checking for verifiable reasoning. Evaluated on the EduMC-QA dataset, MCFR outperformed state-of-the-art LLMs in tasks requiring logical consistency and temporal verification. Our results highlight the limitations of purely neural methods and the value of hybrid approaches. Future work includes integrating Hoare Logic for local consistency checking, complementing model checking for global verification, and enabling scalable, semi-automated model development.


\begin{acks}
We acknowledge Ho Chi Minh City University of Technology (HCMUT), VNU-HCM for supporting this study.
\end{acks}

\vfill\eject
\bibliographystyle{ACM-Reference-Format}
\bibliography{paper_base}

\section*{Appendix: Use of Generative AI Tools}
During the preparation of this paper, we used ChatGPT (developed by OpenAI) solely for the purpose of correcting grammar, spelling, and improving language clarity. The goal of this usage was to enhance readability and ensure linguistic quality, in alignment with academic writing standards. No generative AI tools were used to create research content, perform data analysis, or generate results. All content and ideas presented in the paper are solely the authors' own, and the authors take full responsibility for the final manuscript.


\appendix
\onecolumn
\section{EduMC-QA Knowledge Graph}
\label{app:edu_kg}
\begin{figure*}[h]
    \centering
    \includegraphics[width=0.75\linewidth]{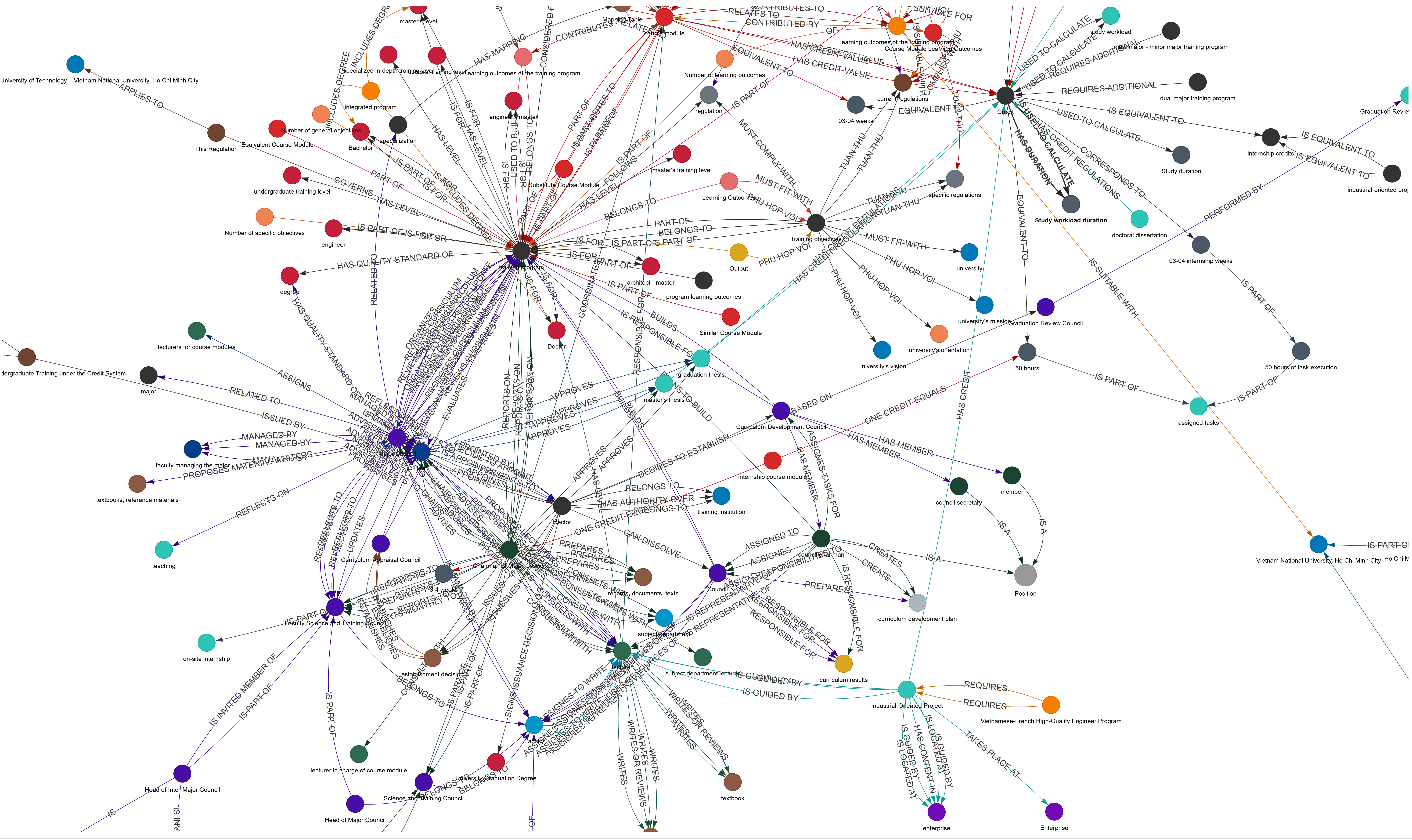}
    \caption{A subgraph of the EduMC-QA Knowledge Graph used in reasoning tasks.}
    \label{fig:edu_kg}
\end{figure*}

\section{Large Language Models Response}
\footnotesize
\label{app:llms_response}
\begin{longtable}{@{} m{3cm} >{\raggedright\arraybackslash}m{1.2cm} >{\raggedright\arraybackslash}m{12.6cm} @{}}
\caption{Comparison of Model Answers and Explanations} \label{tab:model-comparison}\\
\toprule[.1em]
\textbf{Question} & \textbf{Model} & \textbf{Answer \& Explanation} \\
\midrule[.1em]
\endfirsthead

\multirow{4}{*}{\parbox{3cm}{Can a student graduate within eight semesters if all requirements are satisfied on time?}}
& UPPAAL & \textbf{Answer:} Yes. \ding{51} \\
& & \textbf{Explanation:} The UPPAAL query finds a valid trace. This means graduation within eight semesters is feasible.\\
\addlinespace
& ChatGPT & \textbf{Answer:} Yes. \ding{51} \\
& & \textbf{Explanation:} Assuming the program is designed for eight semesters and all requirements are fulfilled (credits, foreign language, IT skills, defense, internship, and thesis), graduation within eight semesters is achievable. \\
\addlinespace
& Claude & \textbf{Answer:} Uncertain. \ding{55} \\
& & \textbf{Explanation:} The context lacks details on credit requirements, standard load per semester, and whether eight semesters is standard. Thus, uncertainty remains. \\
\addlinespace
& Deepseek & \textbf{Answer:} Yes. \ding{51} \\
& & \textbf{Explanation:} If all requirements are met on time, there's no indication of a minimum duration beyond them. Hence, graduation within eight semesters is feasible.\\
\midrule[.1em]

\multirow{4}{*}{\parbox{3cm}{Is it possible to register for CourseX without passing its prerequisite CourseY?}}
& UPPAAL & \textbf{Answer:} No. \ding{51} \\
& & \textbf{Explanation:} Model checking shows no valid path exists, because the guard \texttt{CourseY\_Passed} on the \textsf{CourseX} registration transition blocks all attempts if the prerequisite is not satisfied.\\
\addlinespace
& ChatGPT & \textbf{Answer:} Uncertain. \ding{55} \\
& & \textbf{Explanation:} The context covers retakes and grade improvement but not prerequisite enforcement. Whether Course Y is required before Course X depends on broader academic rules not specified here. \\
\addlinespace
& Claude & \textbf{Answer:} Uncertain. \ding{55} \\
& & \textbf{Explanation:} The context mentions prerequisites for the graduation thesis but provides no general rules on course prerequisites or registration. It doesn't clarify if Course Y must be passed before Course X. \\
\addlinespace
& Deepseek & \textbf{Answer:} Uncertain. \ding{55} \\
& & \textbf{Explanation:} The context doesn't specify rules on course prerequisites or whether they can be bypassed. It mentions retakes and credit accumulation but lacks enough information for a definitive answer. \\
\midrule[.1em]


\multirow{4}{*}{\parbox{3cm}{Can a student advance
to Year 3 within three semesters?}}
& UPPAAL & \textbf{Answer:} No. \ding{51}\\
& & \textbf{Explanation:} The query has no valid path, because completing required courses like \texttt{English\_4} takes at least four semesters.\\
\addlinespace
& ChatGPT & \textbf{Answer:}  Yes. \ding{55}\\
& & \textbf{Explanation:} If a student takes heavy course loads or summer semesters and manages to earn at least 56 credits and complete all additional requirements in 3 semesters, they can advance to Year 3. \\
\addlinespace
& Claude & \textbf{Answer:} Yes. \ding{55}\\
& & \textbf{Explanation:} The context does not specify any minimum time requirements or restrictions that would prevent a student from advancing to Year 3 in 3 semesters, so advancement is possible if all academic requirements are fulfilled. \\
\addlinespace
& Deepseek & \textbf{Answer:} Uncertain. \ding{55}\\
& & \textbf{Explanation:} The context does not provide enough information to determine whether all 3rd-year requirements can be met in 3 semesters. \\
\midrule[.1em]

\multirow{4}{*}{\parbox{3cm}{Could a student accumulate 100 credits yet fail to advance to Year 3 owing to missing social activity points?}}
& UPPAAL & \textbf{Answer:} Yes. \ding{51} \\
& & \textbf{Explanation:} A student can take electives until \texttt{totalCredits}>=100, but keep \texttt{passEnglish\_4}$==false$; the transition to Year 3 is blocked.\\
\addlinespace
& ChatGPT & \textbf{Answer:} No. \ding{55} \\
& & \textbf{Explanation:} The year advancement requirements for Year 3 include credit, language, IT skills, and defense education. Since social activity points aren't listed, lacking them won't block progression. \\
\addlinespace
& Claude & \textbf{Answer:} No. \ding{55} \\
& & \textbf{Explanation:} The context states that Year 3 status requires $\ge$ 56 credits, foreign language, IT skills, and defense education. Social activity points aren't mentioned, so the listed requirements are sufficient for advancement. \\
\addlinespace
& Deepseek & \textbf{Answer:} No. \ding{55} \\
& & \textbf{Explanation:} The requirements for Year 3 include $\ge$ 56 credits, English 1-3, IT skills, and defense education. Social activity points aren't mentioned, so lacking them wouldn't prevent progression if academic conditions are met. \\
\midrule[.1em]

\multirow{4}{*}{\parbox{3cm}{Is it always true that if a student has not passed English 1, then they have not met the second-year foreign language requirement?}}
& UPPAAL & \textbf{Answer:} No. \ding{51}\\
& & \textbf{Explanation:}  Students having IELTS certificates with score above 6 automatically meet all foreign language requirements. \\
\addlinespace
& ChatGPT & \textbf{Answer:} Yes. \ding{55}\\
& & \textbf{Explanation:} According to the regulation, ``Requirements for 2nd-Year Student Status" include ``Passed English 1" as part of the foreign language standard. Therefore, any student who hasn’t passed English 1 clearly does not meet this requirement. \\
\addlinespace
& Claude & \textbf{Answer:} Yes. \ding{55}\\
& & \textbf{Explanation:} According to the regulation, passing English 1 is mandatory for meeting the 2nd-year language standard. As it is the foundation of the sequence, failing it also blocks students from fulfilling the 3rd- and 4th-year standards requiring English 1–3 and 1–4.\\
\addlinespace
& Deepseek & \textbf{Answer:} Yes. \ding{55} \\
& & \textbf{Explanation:} According to the regulation, passing English 1 is required for the 2nd-year language standard. Without it, students cannot meet the requirements for Years 2 to 4.\\
\midrule[.1em]

\multirow{4}{*}{\parbox{3cm}{Can I register for the specialized project if I haven’t completed English 2?}}
& UPPAAL & \textbf{Answer:} Yes. \ding{51}\\
& & \textbf{Explanation: } English 2 is not a requirement for the specialized project.\\
\addlinespace
& ChatGPT & \textbf{Answer:} Yes. \ding{51}\\
& & \textbf{Explanation:} The context does not list English 2 as a requirement for the specialized project. \\
\addlinespace
& Claude & \textbf{Answer:} Uncertain. \ding{55}\\
& & \textbf{Explanation:} The context does not specify any prerequisites for registering for the specialized project.\\
\addlinespace
& Deepseek & \textbf{Answer:} No. \ding{55} \\
& & \textbf{Explanation:} According to the regulations, completing English 2 is an absolute prerequisite before registering for specialized project courses. No exemptions from this requirement are provided in the current rules.\\
\midrule[.1em]

\multirow{4}{*}{\parbox{3cm}{Can I take the internship if I am missing exactly one foundational course?}}
& UPPAAL & \textbf{Answer:} Yes \ding{51}\\
& & \textbf{Explanation:}  Foundational courses are not required to be all complete in take internship. \\
\addlinespace
& ChatGPT & \textbf{Answer:} Uncertain \ding{55}\\
& & \textbf{Explanation:} The context does not mention a specific requirement to complete all foundational courses.\\
\addlinespace
& Claude & \textbf{Answer:} Uncertain \ding{55}\\
& & \textbf{Explanation:} Without knowing the total program credits or how many credits the missing foundational course represents, it is impossible to determine if the student would still meet the 65 \% threshold required for internship eligibility.\\
\addlinespace
& Deepseek & \textbf{Answer:} No. \ding{55}\\
& & \textbf{Explanation:} Students must complete every single foundation course without exception to qualify for internships, as the regulations strictly enforce this prerequisite with no flexibility.\\
\midrule[.1em]

\multirow{4}{*}{\parbox{3cm}{Is it always true that if a student completes English 1–3, they are automatically recognized as having met the Year‑3 foreign language requirement?}}
& UPPAAL & \textbf{Answer:} No. \ding{51}\\
& & \textbf{Explanation:}  Year-3 foreign language requirement is English 1, 2, 3, and 4.\\
\addlinespace
& ChatGPT & \textbf{Answer:} No. \ding{51}\\
& & \textbf{Explanation:} The context clearly states that the Year-3 foreign language requirement is English 1, 2, 3, and 4. Therefore, completing only English 1–3 is not sufficient to meet the requirement.\\
\addlinespace
& Claude & \textbf{Answer:} No. \ding{51}\\
& & \textbf{Explanation:} The context clearly states that the 4th-year foreign language requirement for 3rd-year student status requires "Passed English 1, 2, 3, and 4." A student who has only completed English 1-3 is missing English 4, which is explicitly required. Therefore, completing only English 1-3 does not automatically satisfy the Year-3 foreign language requirement.\\
\addlinespace
& Deepseek & \textbf{Answer:} Yes. \ding{55}\\
& & \textbf{Explanation:} The document stipulates that completing English 1-3 fully satisfies the third-year foreign language requirement. No alternative provisions or additional conditions are mentioned in these regulations.\\

\bottomrule[.1em]
\end{longtable}

\end{document}